\documentclass[letterpaper, 10 pt, conference]{ieeeconf}

\IEEEoverridecommandlockouts
\usepackage{graphicx}
\usepackage{amsmath, amssymb}
\usepackage{url}
\usepackage{cite}
\graphicspath{{fig/}}

\newif\ifanonymous
\anonymousfalse       % camera-ready

\title{\LARGE \bf
PDS Joint: A Parametric Double-Spiral Joint Tailored for Dexterous Hands
}

\author{%
\ifanonymous
Anonymous Authors
\else
Haoyang Li, Yibo Wen, Yixiang Fan, Yiheng Xu, and Yufeng Yue$^{*}$%
\thanks{The authors are with Beijing Institute of Technology, Beijing, China (e-mail: bithaoyangli@163.com).}%
\thanks{$^{*}$Corresponding author: Yufeng Yue (e-mail: yueyufeng@bit.edu.cn).}%
\thanks{Project page: \protect\url{https://spiralhand.github.io/}}%
\fi
}

\begin{document}

\maketitle
\thispagestyle{empty}
\pagestyle{empty}

% -------------------------
% Abstract (replace the current one)
% -------------------------
\begin{abstract}
Compliant joints can embed safety and adaptability into dexterous hands, but achieving large-stroke anthropomorphic motion while maintaining joint-specific, direction-dependent stiffness and reliable proprioception remains challenging.
This paper presents the \emph{PDS joint}, a \textbf{parametric double-spiral (PDS) compliant joint} that enables systematic shaping of directional stiffness across multiple deformation modes, including flexion/extension, abduction/adduction, and pronation/supination.
We instantiate the joint using Archimedean and logarithmic spiral templates for different hand joints and introduce an asymmetry ratio to tailor stiffness distributions for both grasp stability and hyperextension resistance.
To make the joint practically usable under large deformation, we co-design \textbf{embedded inductive proprioception} and propose a \textbf{learning-based calibration pipeline} that maps raw inductive signals to joint states using ArUco-marker tracking.
Experiments characterize the stiffness landscapes across geometric parameters and demonstrate a non-monotonic dependence of lateral support on asymmetry, indicating the importance of principled parameter tuning.
For joint-state estimation in the most challenging abduction/adduction motion, a learned multilayer-perceptron (MLP) mapping reduces the error compared with conventional curve fitting by $41.6\%$.
Finally, we integrate the proposed joints into an open-source dexterous hand as a demonstration platform, on which the hand grasps a set of nine everyday objects and performs safe, contact-rich human-involved interactions.
\end{abstract}

% IROS papers typically omit printed keywords; commented out to match venue style.
% \textbf{Keywords:} compliant joint, dexterous hand, spiral mechanism, proprioception, inductive sensing, calibration

% -------------------------
\section{Introduction}

From simple grippers to dexterous hands, robotic end-effectors are rapidly evolving along with embodied AI.
However, safe and robust deployment in the real world remains a limiting factor, especially in human-robot interaction (HRI), high-speed motion, and cluttered environments, where uncertain impacts and disturbances are inevitable.

Many dexterous hands adopt predominantly stiff joints to achieve high position accuracy.
However, these designs often come with increased distal inertia and tend to amplify the consequence of unexpected collisions or joint over-travel, where hard mechanical stops can cause large impulsive forces on both the robot and its environment.
Moreover, stiff joints often require complicated assembly and rely on accurate sensing and control to maintain performance in contact-rich settings, increasing system complexity.
These challenges motivate joint designs that can inherently tolerate uncertainty and impacts, while remaining accessible to scalable manufacturing and maintenance.

Compliant mechanisms are a compelling alternative because they embed mechanical intelligence into the joint, allowing passive adaptation and the absorption of unexpected impacts.
However, prior compliant hand joints based on elastomers \cite{openhand2013,rus2015soft}, monolithic flexures \cite{naves2017bb}, or rolling contacts \cite{kim2023arc} each trade off repeatability, load-bearing stiffness, or assembly complexity, and are typically dominated by a single bending mode.
This makes it nontrivial to match the varying degrees of freedom (DoF) of human finger joints and limits their applicability beyond simple grippers.

\begin{figure}[t]
  \centering
  \includegraphics[width=\linewidth]{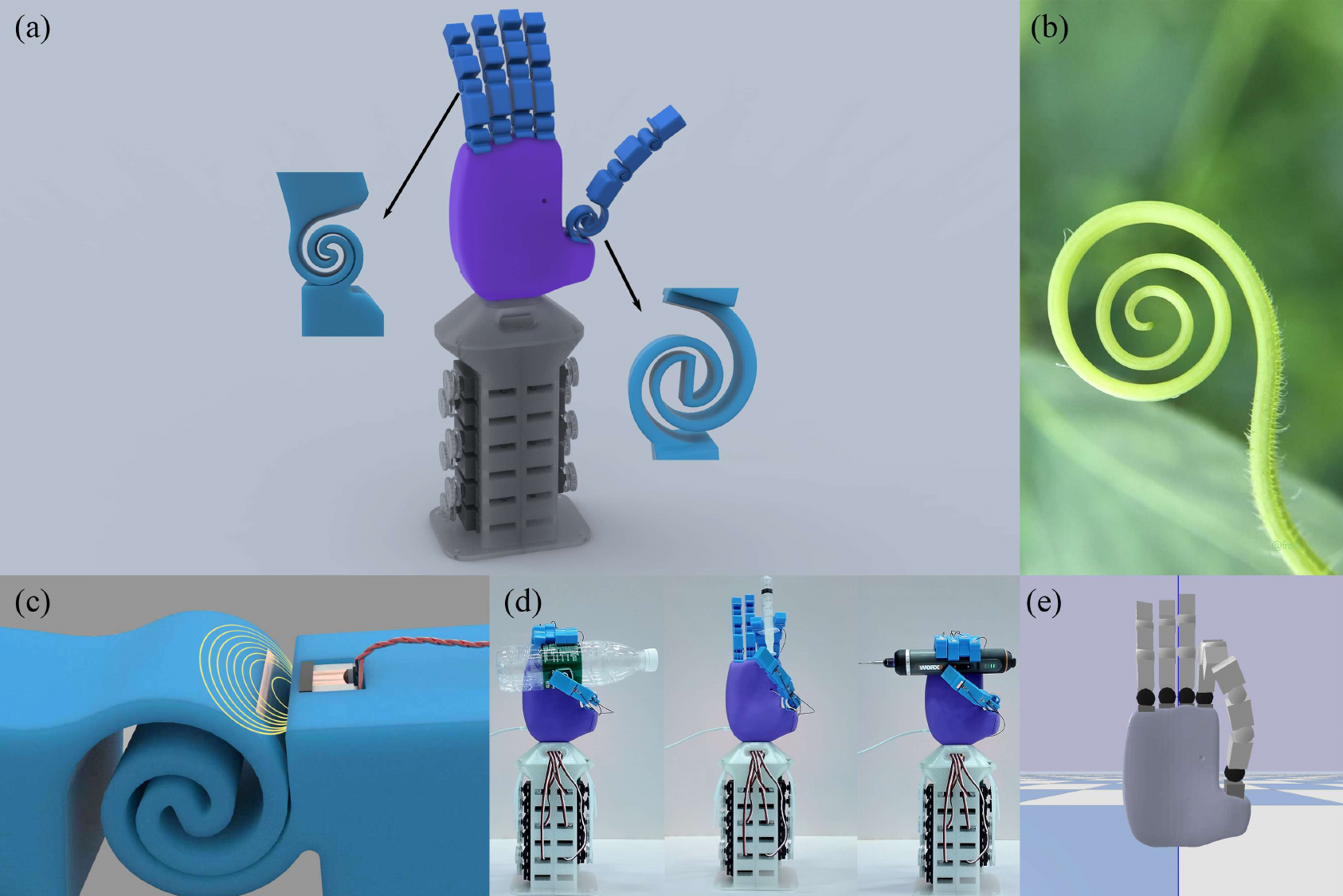}
  \caption{(a) An overview of the PDS-joint-based dexterous hand platform. (b) Demonstration of the spiral morphology in fern fronds. (c) Embedded inductive sensing module. (d) The compliant hand performs grasp and manipulation tasks. (e) A synchronized digital twin enabled by proprioception.}
  \label{fig1}
\end{figure}

Nature offers elegant solutions.
In many fern fronds, tightly coiled fronds provide compact reversible deformation and effectively protect the leaf apical meristem during gradual unfurling \cite{cruz2025fiddlehead,grubb2007adaptive}.
Inspired by such spiral morphologies, spiral-based compliant elements provide a compact geometric template for large-stroke deformation, where mechanical responses can be tuned by a simple set of geometric parameters 
\cite{jafarpour2023doublespiral,jafarpour2024functional}.
However, existing double-spiral studies treat the structure as a general-purpose building block, leaving open how its parameters shape the directional stiffness needed by different hand joints.

Notably, this gap is not purely mechanical.
Complex deformation in compliant finger joints is hard to model accurately, yet recent studies show that, despite their adaptability, compliant hands still require reliable proprioception for complex tasks~\cite{junge2025adapt}.
Among candidate sensing modalities, each of which has distinct limitations for compliant-joint integration~\cite{suzumori2025review}, we adopt inductive sensing, which is robust to electromagnetic compatibility (EMC) disturbances and stray magnetic fields, and pair it with a practical learning-based calibration pipeline using ArUco markers~\cite{brajon2022hybrid,infineon_inductive,ti_tida010961}.

Building on these observations, we present the \emph{PDS joint} and co-design it with embedded inductive proprioception and a learning-based calibration pipeline, integrating the result into an open-source dexterous hand as a \emph{demonstration platform} (Fig.~\ref{fig1}).

\textbf{The key contributions are:}
\begin{itemize}
  \item \textbf{A parameterized double-spiral compliant joint for dexterous hands} that supports large reversible deformation and multi-mode motion, while enabling joint-specific \emph{directional} stiffness shaping (flexion/extension, abduction/adduction, pronation/supination) through a simple geometric parameter set.
  \item \textbf{A systematic parameterization study} that characterizes how geometric parameters influence directional stiffness distribution, providing practical design guidelines for tailoring different finger joints within a dexterous hand.
  \item \textbf{An integrated inductive proprioception system for the double-spiral joints} that is compact, contactless, and robust to EMC disturbances and stray magnetic fields, suitable for real-world contact-rich manipulation \cite{brajon2022hybrid,infineon_inductive,ti_tida010961}.
  \item \textbf{A practical learning-based calibration pipeline} that maps inductive signals to joint states under large deformation, allowing accessible and reliable state estimation for pinch grasping and contact-rich tasks~\cite{junge2025adapt}.
\end{itemize}

% -------------------------
\section{Related Work}

\subsection{Compliant joints for anthropomorphic hands}
The OpenHand family\cite{openhand2013} adopts a hybrid deposition manufacturing (HDM) process that improves robustness and impact resistance, but its compliant hinge exhibits viscoelastic drift and hysteresis that hinder repeatable behavior and state estimation across time and temperature \cite{rus2015soft}.

For large-stroke flexure joints, Naves \emph{et al.} and Garcia \emph{et al.} show that support stiffness in load-bearing directions drops markedly with deflection, identifying off-axis stability as a primary bottleneck that must be addressed by topology design \cite{naves2017bb,garcia2018flexure}.

Rolling-contact joints pursue large-range anthropomorphic kinematics with low friction.  
Kim \emph{et al.} proposed the ARC Joint, which utilizes tongue-and-groove contact with collateral-ligament-like elastic elements to obtain sufficient torsional stiffness for pinch stability \cite{kim2023arc}. 
However, such joints rely on extra elastic components and tight contact constraints, which complicate assembly and compact integration in multi-fingered hands.

Overall, prior compliant joints focus on a single DoF or lack a generalizable geometric paradigm matched to the anthropomorphic stiffness distribution across the hand, motivating a systematic parameterized design.

\subsection{Spiral-inspired compliant elements}
Spiral springs have long been used as compact energy-storing elements with large deformation.
Kim \emph{et al.} proposed a dual-spiral-spring actuation system (DSSAS) for wearable robots, leveraging the spiral spring's high compliance and large deformation. \cite{kim2013dssas}.
Radaelli and Herder conducted an exploratory study on large-displacement spiral springs and demonstrated how varying cross-section properties, material orthotropy, and prestress changes the torsional stiffness, providing design insights for stiffness tailoring \cite{radaelli2018spiral}.

Building on spiral templates, the double-spiral geometry couples two spirals into a compliant element with multiple degrees of freedom and tunable behaviors \cite{jafarpour2023doublespiral,jafarpour2024functional}.
While prior studies establish a versatile library of large-stroke compliant elements, they target general-purpose functionalities or single loading directions; what remains missing for dexterous hand joints is a systematic parameter-to-mechanics mapping that enables joint-specific directional stiffness.

\subsection{Proprioception and calibration for compliant joints}
Recent studies suggest that precise tasks benefit from reliable state feedback rather than pure open-loop control~\cite{junge2025adapt}.
A line of work estimates joint behavior from measurements of actuators. For instance, a minimal-sensing flexure gripper combines displacement sensing with force estimation via direct force sensors, series elastic actuator (SEA) deflection, or actuator current models \cite{min_sensing_gripper}. However, this indirect joint state measurement becomes unreliable with the accumulated error through actuation transmission.

Among electrical sensing modalities, resistive sensors suffer from hysteresis and temperature drift, capacitive sensors from parasitic capacitance and electromagnetic interference, magnetic sensors from stray fields of nearby motors, and optical sensors from routing and bending-induced losses in compact multi-joint hands \cite{rus2015soft,suzumori2025review,aksoy2022shielded}.
In contrast, inductive sensing offers compact, contactless measurement with strong EMC tolerance \cite{infineon_inductive,ti_tida010961}, and magnet-free inductive designs further improve robustness to stray magnetic fields, which is attractive for electromagnetically noisy environments \cite{brajon2022hybrid,renesas_inductive}.
While conventional model-based calibration can be sensitive to manufacturing errors and nonlinear deformation, data-driven calibration with external supervision is widely used in robotics. Fiducial markers such as ArUco provide reliable pose measurements for supervision even under partial occlusion \cite{garrido2014aruco}.
These observations motivate the co-design and integration of the joint and the proprioception system, along with a robust calibration pipeline.

% -------------------------
\section{Approach}

\subsection{Overview}
This section details the PDS joint design, the embedded inductive proprioception, the learning-based calibration pipeline, and the integration of these components into an open-source dexterous hand that serves as a demonstration platform.

\subsection{PDS Joint Design}
Depending on degrees of freedom (DoF) and desired dynamics, we employ two spiral curves.

\begin{figure}[t]
  \centering
  \includegraphics[width=0.8\linewidth]{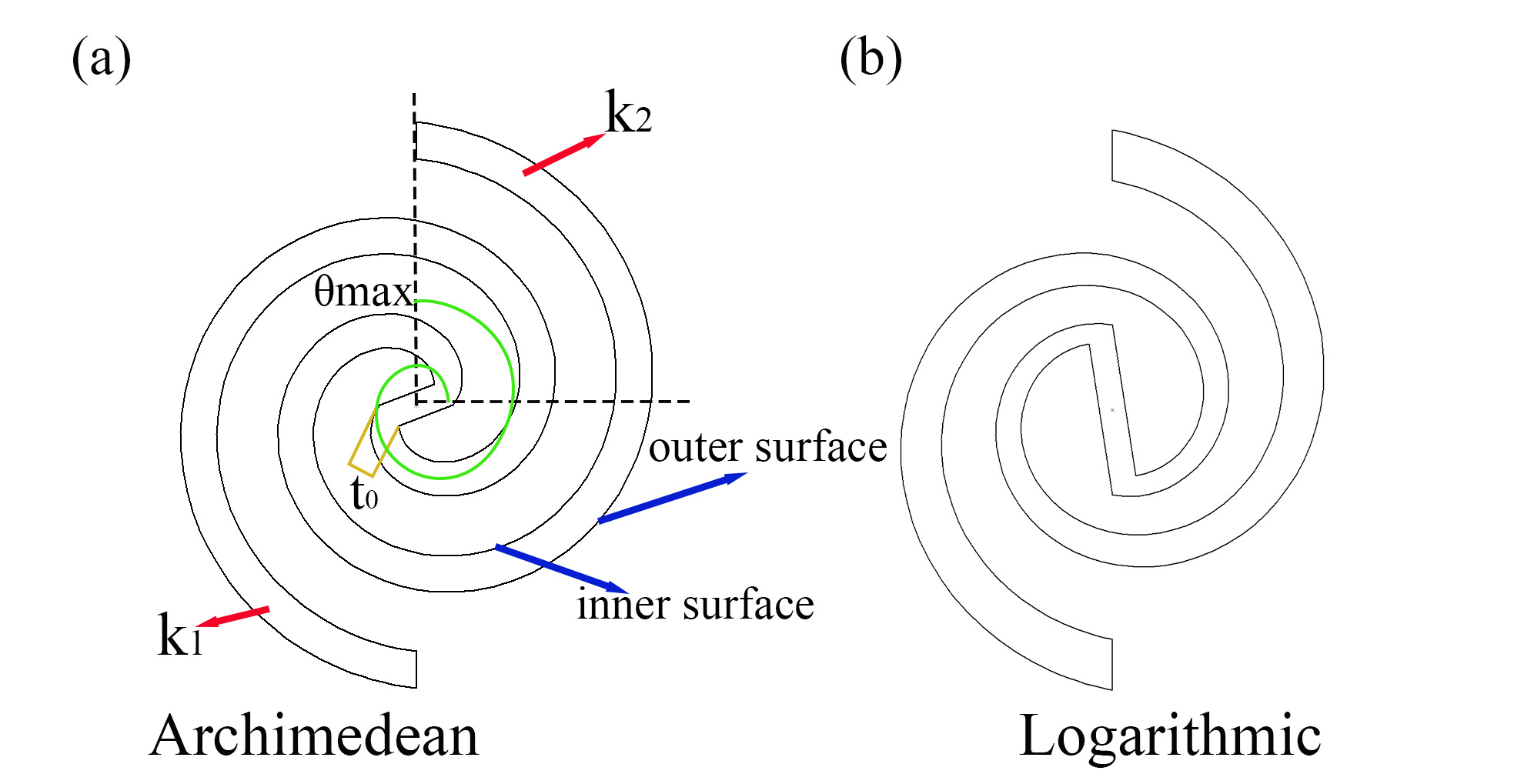}
  \caption{(a)  An Archimedean spiral and the correlating geometric parameters. (b) A logarithmic spiral.}
  \label{fig2}
\end{figure}

\textit{Archimedean spiral.} For the distal interphalangeal (DIP), proximal interphalangeal (PIP), and metacarpophalangeal (MCP) joints of the four fingers, which requires compact design, we use an Archimedean spiral (Fig.~\ref{fig2}):
\begin{equation}
r(\theta) = r_0 + k\theta,\quad \theta \in [0,\theta_{\max}].
\end{equation}

\textit{Logarithmic spiral.} For the carpometacarpal (CMC) joint of the thumb, which requires larger 3D deformation for opposition, we adopt a logarithmic spiral (Fig.~\ref{fig2}):
\begin{equation}
r(\theta) = r_0 e^{k\theta},\quad \theta \in [0,\theta_{\max}].
\end{equation}
Here, $r_0$ is the initial radius, $k$ is the polar slope, and $\theta_{\max}$ defines the angular range.

\begin{figure}[t]
  \centering
  \includegraphics[width=0.8\linewidth]{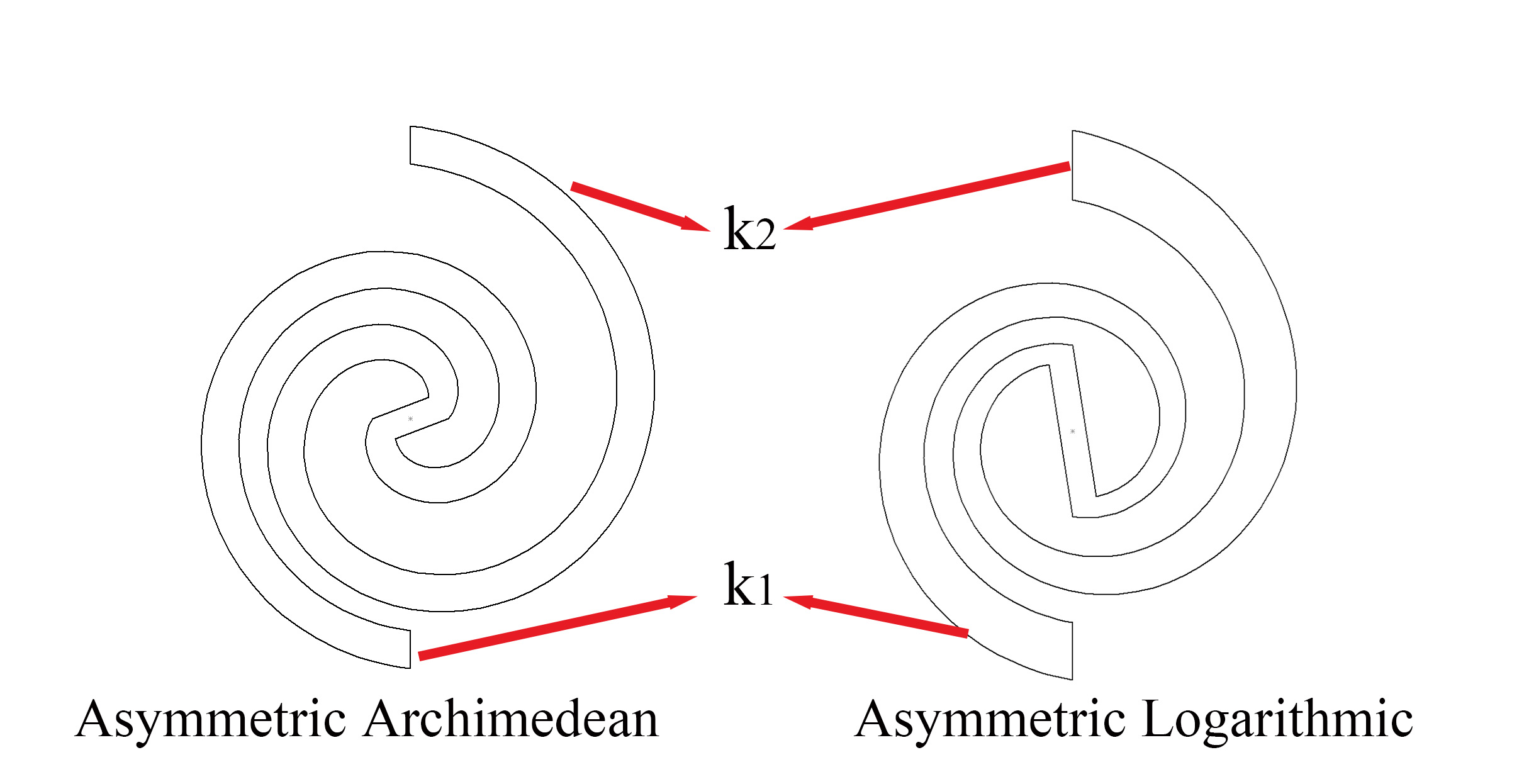}
  \caption{Definition of the asymmetry ratio $\lambda$.}
  \label{fig3}
\end{figure}

The initial thickness is defined as
\begin{equation}
t_0 = r_{0,\text{outer}} - r_{0,\text{inner}}.
\end{equation}

To introduce asymmetry, different slopes are assigned to the two spiral branches, and the asymmetry ratio is defined as
\begin{equation}
\lambda = \frac{k_2}{k_1},
\end{equation}
where $k_1$ and $k_2$ are the polar slopes of opposite branches (Fig.~\ref{fig3}). A symmetric joint corresponds to $\lambda=1$. 

For both types of joint, the geometric parameter set can be concluded as follows.
\begin{equation}
s = \{k_1, k_2, \theta_{\max}, r_{0,\text{inner}}, t_0, \lambda\}.
\end{equation}

The final parameters selected for the hand platform are listed in Table~\ref{tab:joint_params}.

\begin{table}[!b]
\caption{Selected geometric parameters for the final spiral joints.}
\label{tab:joint_params}
\centering
\scriptsize
\begin{tabular}{l c c c c c}
\hline
Joint & Spiral & $\theta_{\max}$ (rad) & $t_0$ (mm) & $k_1$  & $\lambda$ \\
\hline
DIP & Archimedean & $2.48\pi$  & 1.4  & 0.75 & 1.2 \\
PIP & Archimedean & $2.5\pi$  & 1.4  & 0.75 & 1.18 \\
MCP & Archimedean & $2.5\pi$  & 1.4  & 0.8 & 1.22 \\
CMC & Logarithmic & $2\pi$ & 1.5 & 0.18 & 1.12 \\
\hline
\end{tabular}
\end{table}

\subsection{Embedded Inductive Proprioception}
The sensing module includes a customized Printed Circuit Board (PCB), chip inductors, copper foil targets, LDC1614 inductance-to-digital converters~\cite{ti_ldc1614}, and an STM32F405 microcontroller. The sensor exploits eddy-current effects in an LC tank~\cite{peng2024sibs}. As the conductive copper foil approaches the inductor, stronger eddy currents are induced in the target and oppose the coil’s magnetic field, which reduces the coil’s effective inductance. Consequently, the effective inductance $L$ varies nonlinearly but monotonically with the relative distance $d$. This distance-dependent inductance directly modulates the resonant frequency of the LC tank: 
\begin{equation}
f_{\text{res}}(d) = \frac{1}{2\pi\sqrt{L(d)C}},
\end{equation}
where $C$ is the parallel capacitor. 

In practice, the copper foil is attached to the spiral surface and the inductor is embedded in the adjacent phalanx. As the joint bends, the relative distance between the foil and the inductor changes, producing a frequency response correlated to the joint state.

\begin{figure}[t]
  \centering
  \includegraphics[width=\linewidth]{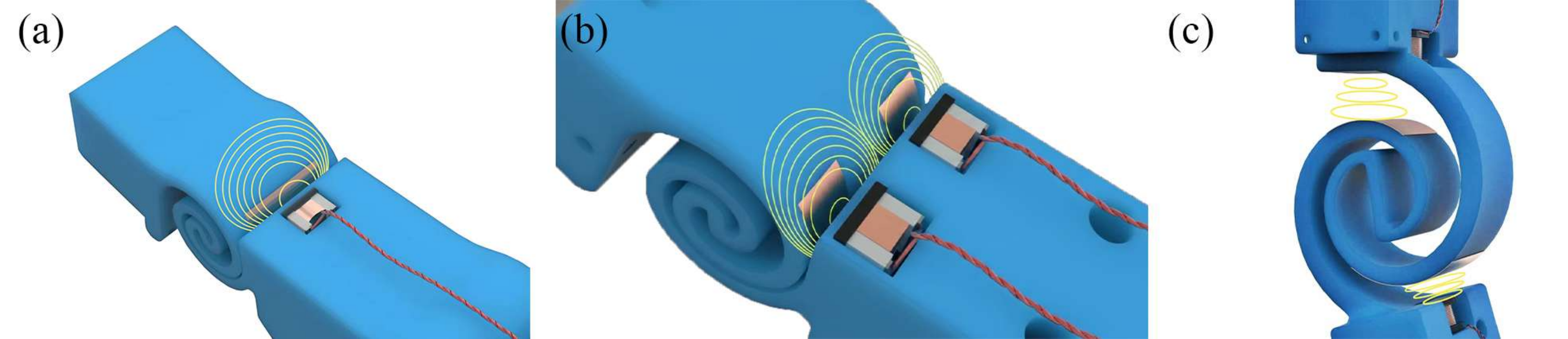}
  \caption{Sensor placement for different joints: DIP/PIP, MCP, and CMC.}
  \label{fig5}
\end{figure}

The sensor placement is tailored to the joint DoF. In particular, there is one channel for DIP/PIP flexion, two channels for MCP to decouple flexion and adduction/abduction via differential signals, and two sensor pairs on opposite sides for the thumb CMC to cover its complex 3D kinematics. The corresponding sensor placements for DIP/PIP, MCP, and thumb CMC are illustrated in Fig.~\ref{fig5}.

\begin{figure}[t]
  \centering
  \includegraphics[width=0.5\linewidth]{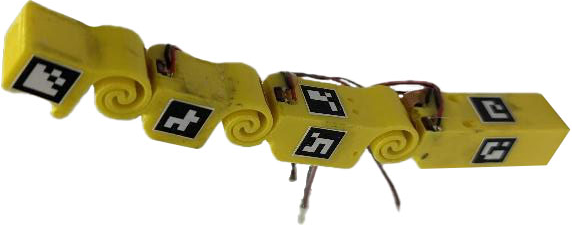}
  \caption{Learning-based calibration using ArUco markers and synchronized sensor acquisition.}
  \label{fig6}
\end{figure}

\subsection{Learning-based Calibration}
To effectively calibrate sensors and map sensor signals to joint states, a calibration pipeline was developed. To start with, sensor signals and joint poses are aligned through three ROS~2 nodes: 
\begin{itemize}
  \item \textit{Pose detection node:} uses a calibrated smartphone camera (30\,FPS) via iVCam to estimate 6D poses through ArUco markers~\cite{garrido2014aruco} pasted on the finger as shown in Fig.~\ref{fig6} with timestamps.
  \item \textit{Signal receiving node:} reads inductive signals from serial (100\,Hz) with timestamps.
  \item \textit{Aligning node:} aligns data streams by timestamps using linear interpolation and saves synchronized data.
\end{itemize}

Next, we establish the mapping between the inductive sensor signals and the joint states. After alignment, we obtain paired samples $\{(\mathbf{z}_t,\mathbf{q}_t)\}_{t=1}^{N}$, where $\mathbf{z}_t$ denotes the inductive sensor output and $\mathbf{q}_t$ denotes the corresponding joint angles measured by the vision pipeline. 

To perform higher-quality mapping, a lightweight multilayer perceptron (MLP) that directly predicts joint states from sensor readings:
\begin{equation}
\hat{\mathbf{q}} = f_{\text{MLP}}(\mathbf{z})
\end{equation}
is preferred. The network is trained to minimize the mean squared error between $\hat{\mathbf{q}}$ and $\mathbf{q}$ in the aligned calibration data in order to better capture coupled nonlinearities and to improve robustness when the mapping deviates from a simple monotonic relationship.

\subsection{Hand Platform Implementation}
To validate the viability of the PDS joint, we developed an open-source 11-DoF dexterous hand platform\footnote{Files are available at \url{https://spiralhand.github.io/}.}. The design details including joint configuration, fingers, palm, and actuation distribution are illustrated below.

\paragraph{Joint configuration}
The joint geometries are defined through the parameter set $\mathbf{s}$. With these parameters, the resulting range of motion (RoM) for each joint is summarized in Table~\ref{tab:rom} and is comparable to that of the human hand.

\begin{table}[t]
\caption{Range of motion (RoM) of Spiral Hand under actuation.}
\label{tab:rom}
\centering
\small
\setlength{\tabcolsep}{6pt}
\begin{tabular}{l l r r r}
\hline
Joint type & Motion & $q_{\min}$ (deg) & $q_{\max}$ (deg) \\
\hline
DIP & Flex./Ext. & -10 & 90 \\
PIP & Flex./Ext. & -15 & 100 \\
MCP & Flex./Ext. & -20 & 90 \\
MCP & Abd./Add.  & -10 & 10 \\
CMC & Flex./Ext. & 0 & 80 \\
CMC & Abd./Add.  & 0 & 90 \\
CMC & Axial rot. & 0 & 90 \\
\hline
\end{tabular}
\end{table}

Notably, the DIP and PIP joints in each finger are driven by a single tendon in an underactuated manner. Although the DIP and PIP joints share a similar RoM in our design, their motion is \emph{sequentially coupled}. During flexion, the PIP joint bends first, followed by the DIP joint, which improves stability during power grasps while maintaining fingertip adaptability.

\begin{figure}[t]
  \centering
  \includegraphics[width=0.65\linewidth]{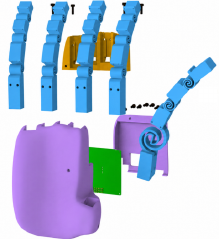}
  \caption{Exploded view of the hand assembly.}
  \label{fig8}
\end{figure}

\paragraph{Modular finger design}
The four non-thumb fingers share an identical module to simplify fabrication and maintenance. Each finger module integrates the PDS joints with the corresponding inductive sensing components, enabling a standardized bill of materials (BoM) and consistent calibration procedures across fingers. This design reduces the diversity of parts and allows damaged modules to be replaced in minutes. An exploded view of the hand assembly is shown in Fig.~\ref{fig8}.

\paragraph{Palm design}
In addition to serving as a mounting base, the geometry of the palm largely determines the grasp envelope, thumb opposition workspace, and overall manipulation capability. Inspired by human hand anatomy, the palm possesses a human-like curvature profile to support stable power grasps and to increase the effective contact area in grasps. The palm structure is designed to accommodate tendon routing and the sensor wiring while keeping the assembly compact and convenient for maintenance.

\paragraph{Efficient actuation distribution}
Rather than fully actuating every joint, we allocate independent actuation to critical DoFs and underactuate others to reduce cost. Specifically, the MCP flexion/extension of index and middle fingers is independently actuated to provide precise control of finger posture during pinch and tool use. The DIP and PIP joints are designed to flex in a coupled manner under a single actuation input, reducing actuation complexity while preserving an effective flexion motion. This coupling also improves robustness in contact by allowing the distal phalanges to self-adjust as they wrap around objects. 

The ring and pinky fingers are designed with complete underactuated flexion. This design reduces actuator count while enabling robust power grasps and enveloping grasps, where the ring and pinky fingers primarily form grasp closure rather than precise fingertip motions.

\begin{figure}[t]
  \centering
  \includegraphics[width=\linewidth]{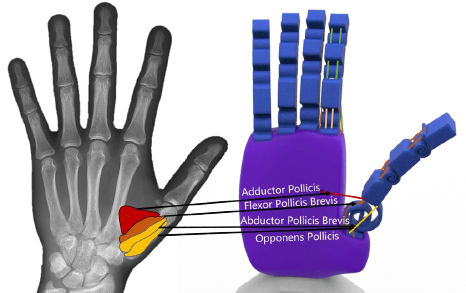}
  \caption{Specialized actuation distribution of the thumb.}
  \label{fig9}
\end{figure}

The actuation of the thumb is specialized to enable complex opposition behaviors (Fig.~\ref{fig9}). We under-actuate the IP, MCP, and CMC joint in flexion, yielding a coordinated motion commonly used in grasp closure. In addition, an independent tendon realizes the function of the adductor pollicis, and another independent tendon combines the function of the opponens pollicis and abductor pollicis brevis. This actuation distribution preserves not only the ability of the thumb to perform power grasps, but also its ability to perform complex manipulation tasks.

% --------------------------
\section{Experiments}
We evaluate the viability of the PDS joint by addressing the following questions:

(Q1) Can the PDS parameters systematically shape the directional stiffness across DoFs? (A)

(Q2) Can embedded inductive sensing provide reliable signals under large deformation? (B)

(Q3) Can learning-based calibration map raw signals to joint states accurately? (C)

(Q4) Do these capabilities enable safe and robust contact-rich demonstrations on a hand platform? (D)

\subsection{Common Setup and Definition}
\paragraph{Samples and fabrication}
All of the samples are 3D printed via Fused Deposition Modeling (FDM) printers under identical settings and share the same Thermoplastic Urethane (TPU 68D) filament. The parameter sets adopted in the samples are summarized in Table~\ref{tab:exp_design}. 

\begin{table}[!b]
\caption{The parameter sets for the experiments.}
\label{tab:exp_design}
\centering
\small
\setlength{\tabcolsep}{5pt}
\begin{tabular}{lccccc}
\hline
ID & $\theta_{\max}$ (rad) & $t_0$ (mm) & $k_1$ & $k_2$ & $\lambda{=}k_2/k_1$ \\
\hline
A1 & $1.5\pi$  & 1.4 & 0.8 & 0.8 & 1.000 \\
A2 & $1.75\pi$ & 1.4 & 0.8 & 0.8 & 1.000 \\
A3 & $2.0\pi$  & 1.4 & 0.8 & 0.8 & 1.000 \\
A4 & $2.25\pi$ & 1.4 & 0.8 & 0.8 & 1.000 \\
A5 & $2.5\pi$  & 1.4 & 0.8 & 0.8 & 1.000 \\
\hline
K1 & $2.5\pi$ & 1.4 & 0.6 & 0.6 & 1.000 \\
K2 & $2.5\pi$ & 1.4 & 0.7 & 0.7 & 1.000 \\
K3 & $2.5\pi$ & 1.4 & 0.8 & 0.8 & 1.000 \\
K4 & $2.5\pi$ & 1.4 & 0.9 & 0.9 & 1.000 \\
K5 & $2.5\pi$ & 1.4 & 1.0 & 1.0 & 1.000 \\
\hline
T1 & $2.5\pi$ & 0.8 & 0.8 & 0.8 & 1.000 \\
T2 & $2.5\pi$ & 1.0 & 0.8 & 0.8 & 1.000 \\
T3 & $2.5\pi$ & 1.2 & 0.8 & 0.8 & 1.000 \\
T4 & $2.5\pi$ & 1.4 & 0.8 & 0.8 & 1.000 \\
T5 & $2.5\pi$ & 1.6 & 0.8 & 0.8 & 1.000 \\
\hline
R1 & $2.5\pi$ & 1.4 & 0.8 & 0.70 & 0.875 \\
R2 & $2.5\pi$ & 1.4 & 0.8 & 0.75 & 0.938 \\
R3 & $2.5\pi$ & 1.4 & 0.8 & 0.80 & 1.000 \\
R4 & $2.5\pi$ & 1.4 & 0.8 & 0.85 & 1.062 \\
R5 & $2.5\pi$ & 1.4 & 0.8 & 0.90 & 1.125 \\
\hline
\end{tabular}
\end{table}

\paragraph{DoF and coordinate definitions}
We refer to the DoFs of the joints as: flexion/extension (FE), abduction/adduction (AA), and pronation/supination (PS). The flexion, abduction, and pronation is defined as the positive direction of motion. All of the zero-positions refer to the initial state of the printed joints unless otherwise noted.

\paragraph{Angle and torque measurement}
Joint angles were measured using ArUco-based pose estimation at 30~Hz.
For stiffness characterization, we applied joint torques using a constant moment-arm mechanism. The resulting forces were measured with a high-resolution dynamometer, which was motor-driven to maintain a constant force unless otherwise noted.

\subsection{Mechanical Characterization}

\begin{figure}[t]
  \centering
  \includegraphics[width=\linewidth]{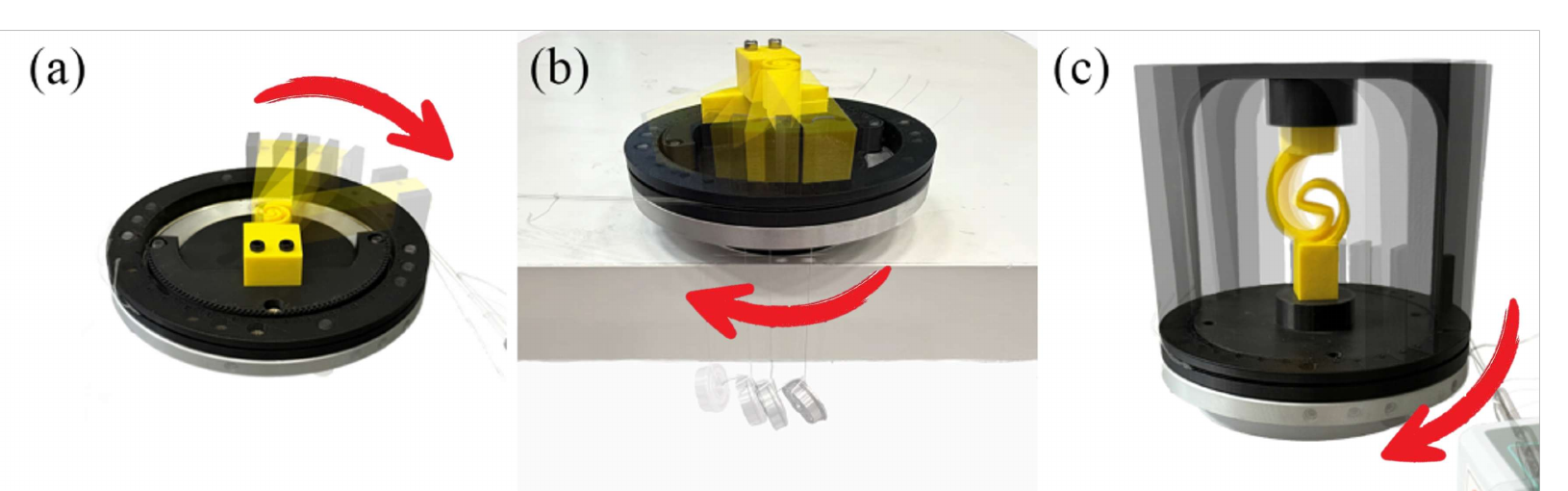}
  \caption{(a) Experimental setup for mechanical characterization in flexion/extension. (b) Experimental setup for mechanical characterization in abduction/adduction (c) Experimental setup for mechanical characterization in pronation/supination.}
  \label{fig10}
\end{figure}

\subsubsection{Flexion/Extension Stiffness Characterization}
\paragraph{Protocol}
The torque-angle relationship of the PDS joints during flexion was measured with the setup shown in Fig.~\ref{fig10}. Notably, measuring the flexion characterization of a PDS joint with the asymmetry ratio $\lambda$ is comparable to the extension characterization of the joint with $\lambda' = 1/\lambda$. 

\begin{figure}[t]
  \centering
  \includegraphics[width=\linewidth]{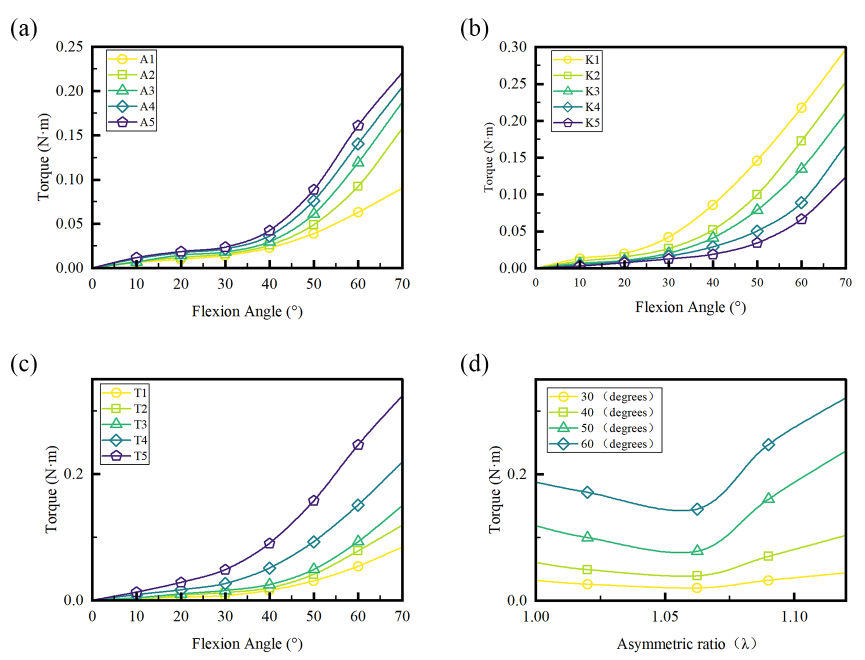}
  \caption{(a) Measured flexion/extension torque-angle curves with varying $\theta_{\max}$. (b) Measured flexion/extension torque-angle curves with varying k. (c) Measured flexion/extension torque-angle curves with varying $t_0$. (d) Processed flexion/extension torque-angle curves with varying $\lambda$.}
  \label{fig11}
\end{figure}

\paragraph{Results}
As shown in Fig.~\ref{fig11}, increasing $\theta_{\max}$ and $t_0$ shifts the torque profiles upward, while increasing slope $k$ tends to reduce stiffness due to larger gaps between the coils. In particular, the asymmetry ratio $\lambda$ shows a non-monotonic relationship with the actuating torque: when $\lambda>1$, the torque may decrease slightly before rising rapidly, with a local minimum observed near $\lambda \approx 1.06$ in our measurements.

\subsubsection{Abduction/Adduction Stiffness Characterization}
\paragraph{Protocol}
The lateral-stiffness-angle relationship of the PDS joint was measured under constant lateral force (0.5N) from a hanging weight at varying flexion angles, using the setup shown in Fig.~\ref{fig10}.

\begin{figure}[t]
  \centering
  \includegraphics[width=\linewidth]{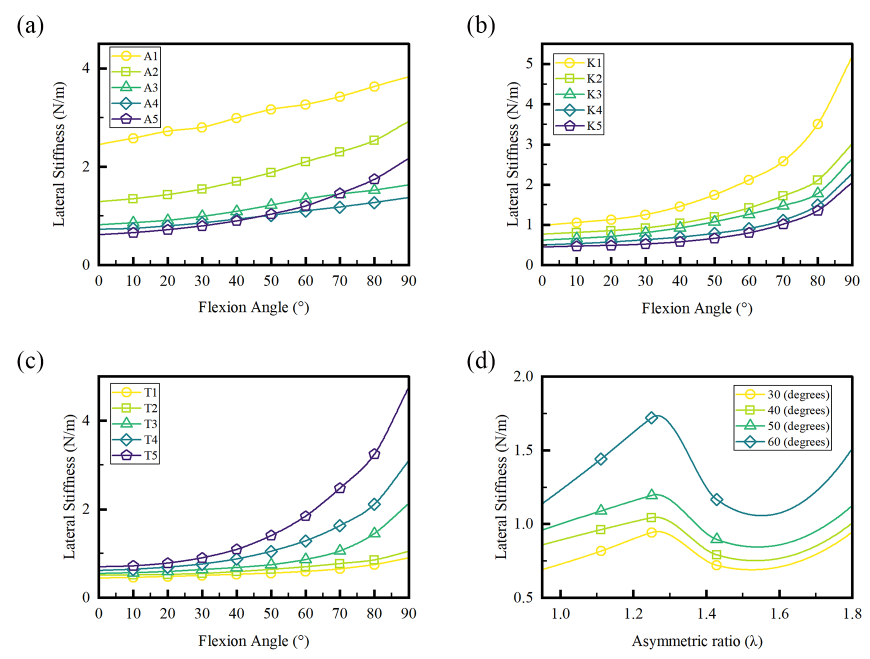}
  \caption{(a) Measured lateral stiffness versus flexion angle with varying $\theta_{\max}$. (b) Measured lateral stiffness versus flexion angle with varying k. (c) Measured lateral stiffness versus flexion angle with varying $t_0$. (d) Processed lateral stiffness versus flexion angle with varying $\lambda$.}
  \label{fig12}
\end{figure}

\paragraph{Results}
Fig.~\ref{fig12} shows that in all samples, the lateral stiffness increases with the flexion angle. The profiles shift downward as $\theta_{\max}$ and $k$ increase, while increasing $t_0$ shifts the profiles upward. When $\theta_{\max}$ increases to $2.5\pi$, the lateral stiffness increases dramatically with increasing flexion, which is beneficial to provide compliance during the approach while maintaining rigidity in grasp. Finally, as the asymmetry ratio $\lambda$ increases over 1.00, the lateral stiffness first increases, reaching a maximum at around $\lambda = 1.25$, and then presents another non-monotonic behavior, with a minimum at approximately $\lambda = 1.55$. This nontrivial relationship indicates that the asymmetric design effectively improves the lateral stiffness in a specific range, and highlights the importance of fine parameter tuning rather than simply maximizing $\lambda$.

\subsubsection{Pronation/Supination Stiffness Characterization}
\paragraph{Protocol}
Measurements of torsional torque versus twist angle for the thumb CMC joint were conducted with the setup illustrated in Fig.~\ref{fig10}.

\begin{figure}[t]
  \centering
  \includegraphics[width=\linewidth]{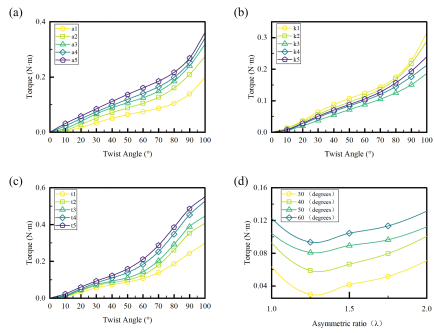}
  \caption{(a) Measured pronation/supination torque-angle curves with varying $\theta_{\max}$. (b) Measured pronation/supination torque-angle curves with varying k. (c) Measured pronation/supination torque-angle curves with varying $t_0$. (d) Processed pronation/supination torque-angle curves with varying $\lambda$.}
  \label{fig13}
\end{figure}

\paragraph{Results}
Measured torque correlates positively with twist angle, as shown in Fig.~\ref{fig13}. Increasing $\theta_{\max}$ and $t_0$ shifts the profiles upward, while $k$ exhibits a non-monotonic effect. A similar non-monotonic trend of torque profiles is observed with respect to $\lambda$ to that in the flexion/extension experiment. The local minimum appears around $\lambda \approx 1.26$ in this case.

Based on these findings, we can decide the parameters for the final hand design. For DIP and PIP joints, we pursue high lateral stiffness, while for the MCP joint, reduced lateral stiffness is preferred to enable easier adduction/abduction. For the CMC joint, we prioritize easier axial rotation and lower lateral stiffness for opposition.

\subsubsection{Repeatability Under Cyclic Loading}
\paragraph{Protocol}
To evaluate long-term repeatability under repeated large deformation, we conducted a cyclic loading test on the PDS joint.
Each cycle consisted of flexing the joint from its natural position to its maximum flexion angle, i.e., the rated flexion limit in Table~\ref{tab:rom}, and releasing the load to observe the drift angle from zero-position.
The joint was actuated at a frequency of 2~Hz, and the test was repeated for 2000 cycles under identical conditions at room temperature.

\begin{figure}[t]
  \centering
  \includegraphics[width=0.65\linewidth]{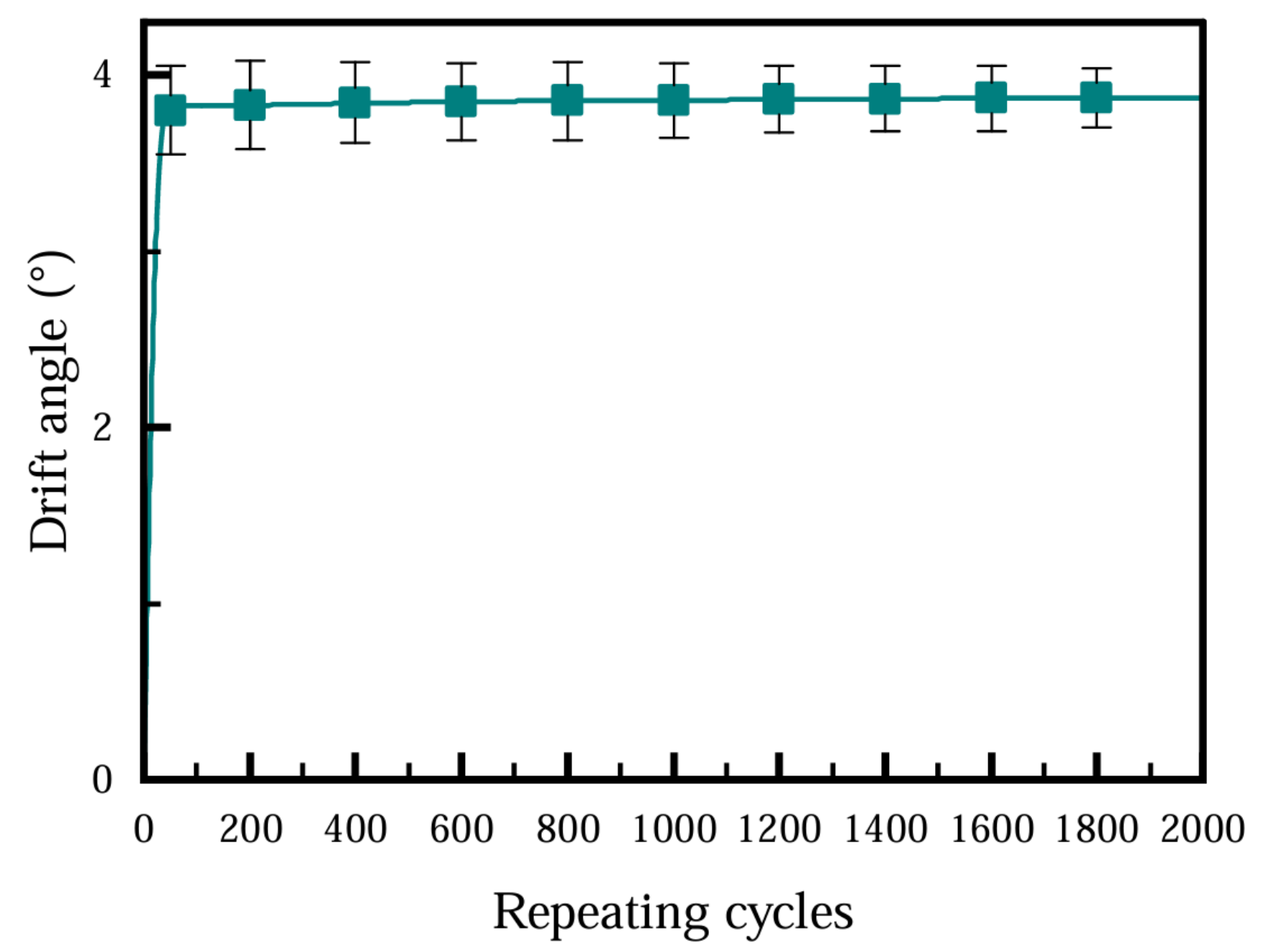}
  \caption{Drift angle versus cycle count in the cyclic loading test.}
  \label{fig15}
\end{figure}

\paragraph{Results}
As shown in Fig.~\ref{fig15}, the drift angle quickly increases to around $3.8^\circ$, and then remained at a relatively stable offset from zero-position. Based on this observation, we deliberately enlarged the extension angle at zero-position slightly to enable the desired RoM.

\subsection{Proprioception and Calibration Evaluation}

\subsubsection{Raw Inductive Signal Characterization}
\begin{figure}[t]
  \centering
  \includegraphics[width=\linewidth]{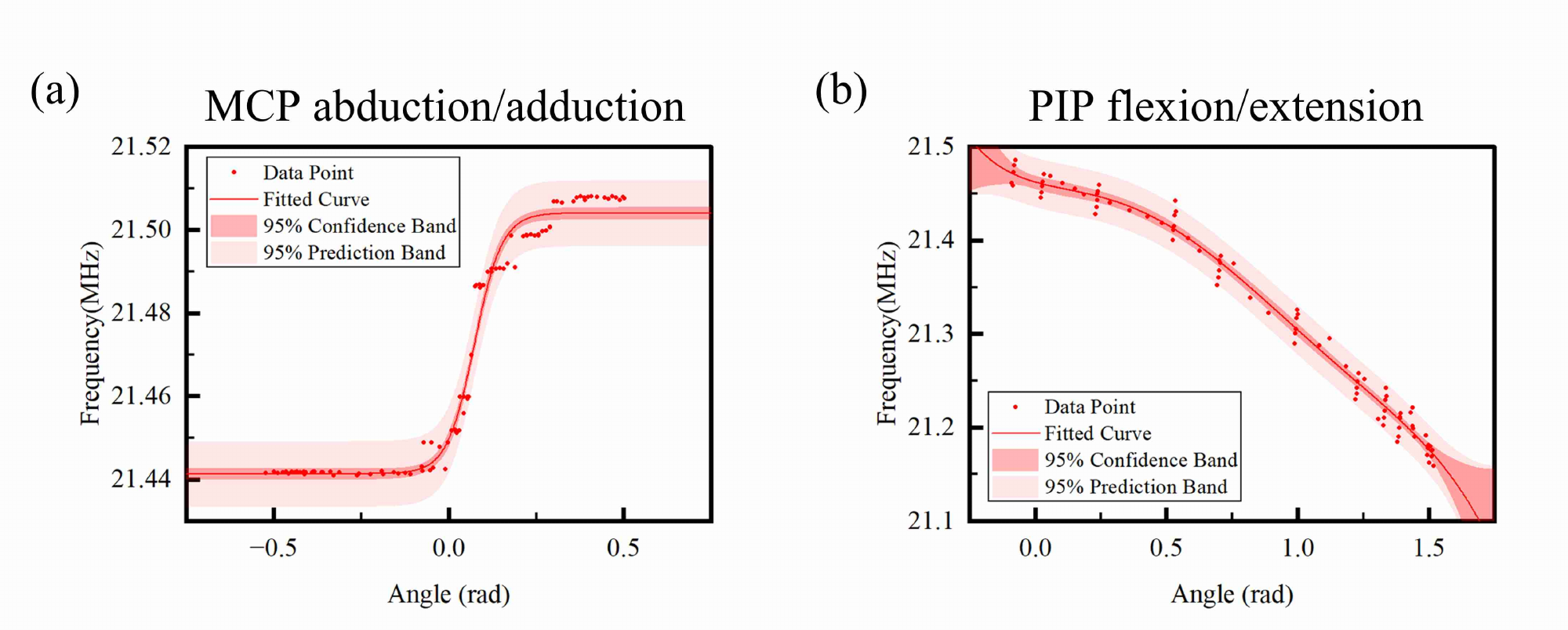}
  \caption{Raw signal characterization: fitted curves of raw inductive signal versus joint angle. (a) MCP abduction/adduction; (b) PIP flexion/extension.}
  \label{fig16}
\end{figure}

\paragraph{Protocol}
To illustrate the effectiveness of the inductive sensing system, the raw-signal--bending-angle relationships of the PIP and MCP joints are presented in Fig.~\ref{fig16}.

\paragraph{Results}
The single-DoF PIP joint demonstrated an approximately linear relationship between the signal and the joint angle. However, in AA motion of MCP joints, where the distance $d$ between the inductor and the copper target is significantly narrowed, signal saturation is observed at large angles.

\subsubsection{Joint-State Mapping: Fitting vs. MLP}
\paragraph{Protocol}
To evaluate the effectiveness of the MLP calibration pipeline, we present the parity plots of ground truth angles and estimated angles in the most difficult AA motion. Then, we compare the learning-based method with conventional curve fitting method through the plots.

\begin{figure}[t]
  \centering
  \includegraphics[width=\linewidth]{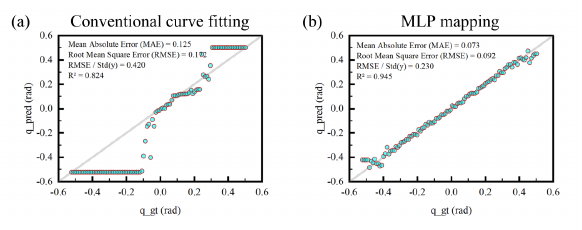}
  \caption{Joint-state mapping in the challenging MCP abduction/adduction case: (a) conventional fitting versus (b) MLP-based method.}
  \label{fig17}
\end{figure}

\paragraph{Results}
As Fig.~\ref{fig17} suggests, the MLP method achieves a mean absolute error (MAE) of 0.073 and a root-mean-square error (RMSE) of 0.092, outperforming the conventional curve fitting method (MAE = 0.125, RMSE = 0.172).

\subsection{Demonstration on Hand Platform}

\subsubsection{Adaptive grasping}
\begin{figure}[t]
  \centering
  \includegraphics[width=0.75\linewidth]{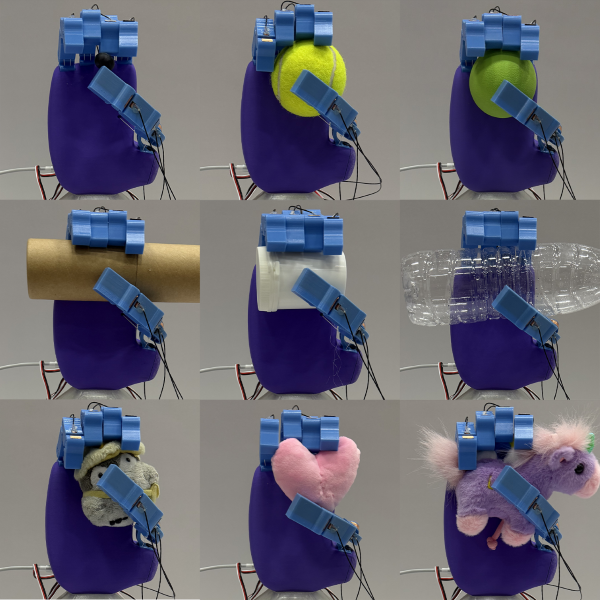}
  \caption{Adaptive grasping demonstrations across spherical, cylindrical, and irregular/deformable objects.}
  \label{fig18}
\end{figure}

\paragraph{Protocol}
We evaluate grasping robustness using the hand platform equipped with PDS joints.
A set of nine everyday objects spanning three categories was used: spherical objects whose mass ranges from a few grams (a blueberry) to about $300$\,g (a fascia ball), cylindrical objects, and irregular or deformable objects such as stuffed toys, as shown in Fig.~\ref{fig18}.
For each trial, the hand starts from a fixed initial pose and closes the fingers under the same position sequence command. 
The demonstrations rely on the inductive proprioception system to observe joint states during grasps.
Each object is tested for 50 trials, and a stable grasp maintained over 30 seconds is considered successful.

\paragraph{Results}
The compliant hand, integrated with the PDS joints and the inductive sensing system, grasped all nine objects successfully across all trials, with representative poses shown in Fig.~\ref{fig18}. 

\begin{figure}[t]
  \centering
  \includegraphics[width=0.75\linewidth]{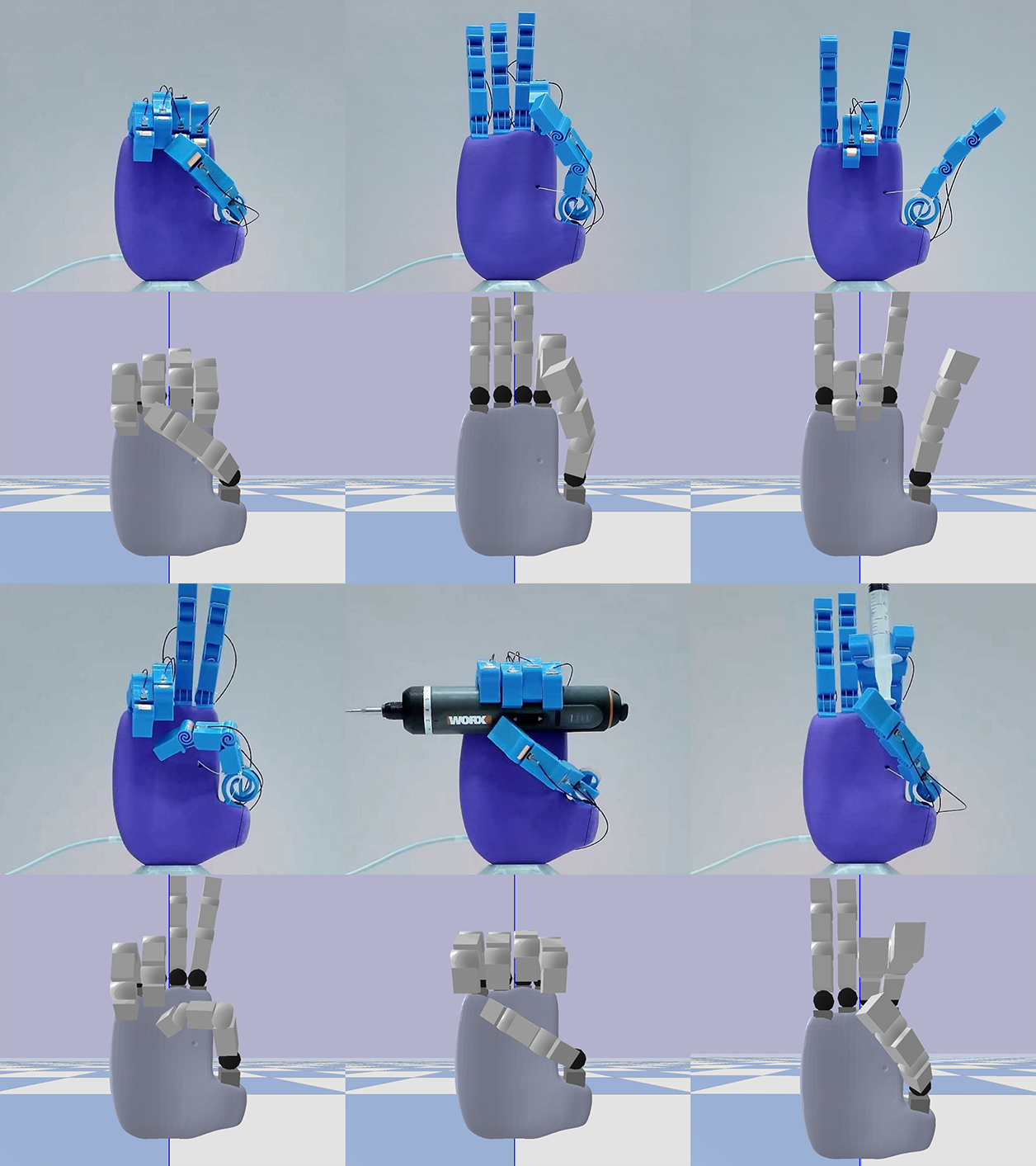}
  \caption{Contact-rich manipulation demonstrations and synchronized PyBullet visualization.}
  \label{fig19}
\end{figure}

\subsubsection{Contact-rich Demonstrations with Safety}
\paragraph{Protocol}
To further investigate the effectiveness of the design, we implemented the Kapandji test, and two contact-rich manipulation tasks involving pinch grasp and thumb opposition. Firstly, we placed a syringe between the index and the middle fingers in motion, and let the hand pinch grasp it and push the plunger upward. Next, we let the hand take hold of an electric screwdriver in motion and press the button to drive it. In addition, the safety of compliant joints is intuitively illustrated in these tasks.

\begin{figure}[t]
  \centering
  \includegraphics[width=0.75\linewidth]{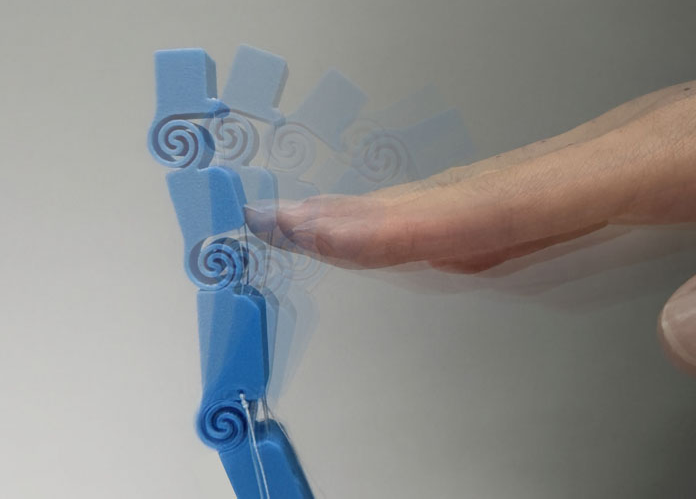}
  \caption{Illustration of impact absorption and safe human-involved interaction enabled by the compliant PDS joint.}
  \label{fig20}
\end{figure}

\paragraph{Results}
Our hand platform scored 10/10 in the Kapandji test, and successfully performed these tasks as shown in Fig.~\ref{fig19}, demonstrating the dexterity provided by the PDS joint. Fig.~\ref{fig20} illustrates how the PDS joint absorbs impact and collision in use, ensuring safety in human-involved tasks.

% --------------------------
\section{Conclusion}
This paper presented the \emph{PDS joint}, a parametric double-spiral compliant joint designed for dexterous hands.
By parameterizing double-spiral geometries, the proposed design provides a unified geometric paradigm to tailor stiffness distributions across flexion/extension, abduction/adduction, and pronation/supination, enabling both large-stroke motion and improved off-axis support within selected parameter ranges.
To address the state-estimation challenge inherent in compliant mechanisms, we integrated compact inductive proprioception into the joint and introduced a practical learning-based calibration pipeline using ArUco-supervised data, which improved joint-state mapping accuracy over conventional curve fitting.

Comprehensive experiments validated the approach from component to system levels.
We investigated the stiffness characteristics over key geometric parameters and demonstrated repeatability under cyclic loading.
The inductive sensing signals exhibited repeatable trends with joint configuration, and the learning-based calibration achieved lower estimation error in challenging motions.
Finally, integration on an open-source dexterous hand demonstrated safe and robust contact-rich tasks, including adaptive grasping and human-involved tool operations.

This study is demonstration-oriented and has clear limitations that define our future work. Grasping was evaluated on a limited set of light, regular objects without baseline comparisons against rigid joints, non-perceptive variants, or alternative compliant joints, and the benefit of proprioception for closed-loop manipulation was not yet quantified; likewise, the cyclic-loading test captures only short-term drift rather than long-term fatigue. Future work will therefore add quantitative baselines and longer fatigue evaluations under broader loading conditions, improve sensor placement to mitigate saturation at extreme poses, and leverage the calibrated proprioception for closed-loop manipulation and higher-level policies such as imitation learning (IL) and reinforcement learning (RL).

% -------------------------
% Double-blind note:
% Keep acknowledgments and identifying links removed for anonymous submission.
% -------------------------
% \section*{ACKNOWLEDGMENTS}
% (Uncomment for camera-ready.)

% -------------------------
% References (placeholder)
% -------------------------
\bibliographystyle{IEEEtran}
\bibliography{refs}

\end{document}